\documentclass[conference]{IEEEtran}

\IEEEoverridecommandlockouts                              

\title{\LARGE \bf
GI-NNet \& RGI-NNet: Development of Robotic Grasp Pose Models, Trainable with  Large as well as Limited Labelled Training Datasets, under supervised and semi supervised paradigms.
}

\author{Priya Shukla$^{1}$, 
Nilotpal Pramanik$^{1}$, Deepesh Mehta$^{2}$ and G.C. Nandi$^{3}$
\thanks{$^{1}$Student at Center of Intelligent Robotics, Indian Institute of Information Technology, Allahabad, Prayagraj, India-211015 
        {\tt\small priyashuklalko@gmail.com}}%
\thanks{$^{2}$Winter intern at Center of Intelligent Robotics, Indian Institute of Information Technology, Allahabad, Prayagraj, India-211015}
\thanks{$^{3}$Professor at Center of Intelligent Robotics, Indian Institute of Information Technology, Allahabad, Prayagraj, India-211015
        {\tt\small gcnandi@iiita.ac.in}}%
}
\usepackage{caption}
\usepackage{array}
\usepackage{tabularx}
\usepackage{amsmath}
\usepackage{graphicx}
\usepackage{gensymb}
\usepackage{lipsum}
\usepackage{xcolor}
\graphicspath{ {images/} }
\usepackage[export]{adjustbox}
\usepackage{hyperref}
\usepackage{adjustbox}
\usepackage{romannum}
\usepackage{comment}

\begin{document}

\maketitle
\pagenumbering{arabic}
\pagestyle{plain}

\begin{abstract}
Grasp manipulations by  COBOTS (COllaborative roBOTS) are supposed to be functioning the same way we grasp objects,  even complex objects. However, for robots executing an intelligent and optimal grasp efficiently, the way we grasp objects, is quite challenging. The reason being that we acquire this skill by spending  a lot of time in our childhood trying and failing to pick things up, and learning from our mistakes. For robots we can’t wait through the equivalent of an entire robotic childhood. To streamline the process, in the present investigation we propose to use deep learning techniques to help robot learn quickly how to generate and execute appropriate grasps. More specifically, we develop two models. One   a Generative Inception Neural Network (GI-NNet) model, capable of generating antipodal robotic grasps on seen as well as unseen objects. It is trained on Cornell Grasping Dataset (CGD) and performed excellently by attaining 98.87\% grasp pose accuracy by detecting the same from the RGB-Depth (RGB-D) images for regular as well as irregular shaped objects while it requires only one third of the network trainable parameters as compared to the State-Of-The-Art (SOTA) approach \cite{kumra2019antipodal}.
However, to attain this level of performance the model requires the entire 90\% of the available labelled data of CGD keeping only remaining 10\% labelled data for testing, making it vulnerable for the poor generalization. Also, as we all know, getting sufficient and quality labelled dataset is becoming increasingly difficult keeping in pace with the requirement of gigantic networks. To address these issues, we subsequently propose to attach our model as decoder with a semi-supervised learning based architecture known as Vector Quantized Variational Auto Encoder (VQVAE), which we make to work efficiently when we train it with available labelled dataset as well as unlabelled data \cite{mridul}.  Our proposed GI-NNet integrated   VQVAE model, which we name as Representation based GI-NNet (RGI-NNet), has been trained with various splits of label data on CGD with as minimum as 10\% labelled dataset together with latent embedding generated from VQVAE up to 50\% labelled data with latent embedding obtained  from VQVAE. The performance level, in terms of grasp pose accuracy of RGI-NNet, varies between 92.13\% to 95.6\% which is far better than many other existing SOTA models trained with only labelled dataset. For the performance verification of both GI-NNet and RGI-NNet models, we use  Anukul (Baxter) hardware  cobot and it is observed that both the proposed models performed significantly better in real time table top grasp executions.
The logical details of our proposed models together with an in depth analyses have been presented in the paper.
\end{abstract}


\section{Introduction}
\label{intro}
Recently, with the advancement of deep learning technologies, the capability of robots are enhancing day by day from manipulating very rudimentary type of tasks such as palletizing, picking/placing \cite{bohg2013data,c11} to the complicated tasks like trying to grasp and manipulate previously seen as well as unseen objects intelligently, the way we do, so that robots can share work space with us in a social environment including many household applications. However, to make it happen, we need to solve many challenging problems one of them is to make it learn the tasks it requires to perform like human kids learn over the years. A child normally has poor grasping skill and that's the reason we are reluctant to  allow them to grasp sophisticated items for fear of damaging them. But the same child when grew up develops enough grasping skills based on learning through experience, the mechanism of which we need to know before we apply similar mechanisms to train robots including social robots so that they can manipulate tasks more adaptively in a realistic household and industrial environment. 
As expanding research is being continued to make the robots more intelligent, there exists an interest for a summed up method to induce quick and powerful grasps for any sort of item that the robot sees.

\begin{figure}
\centering
\includegraphics[scale=0.25]{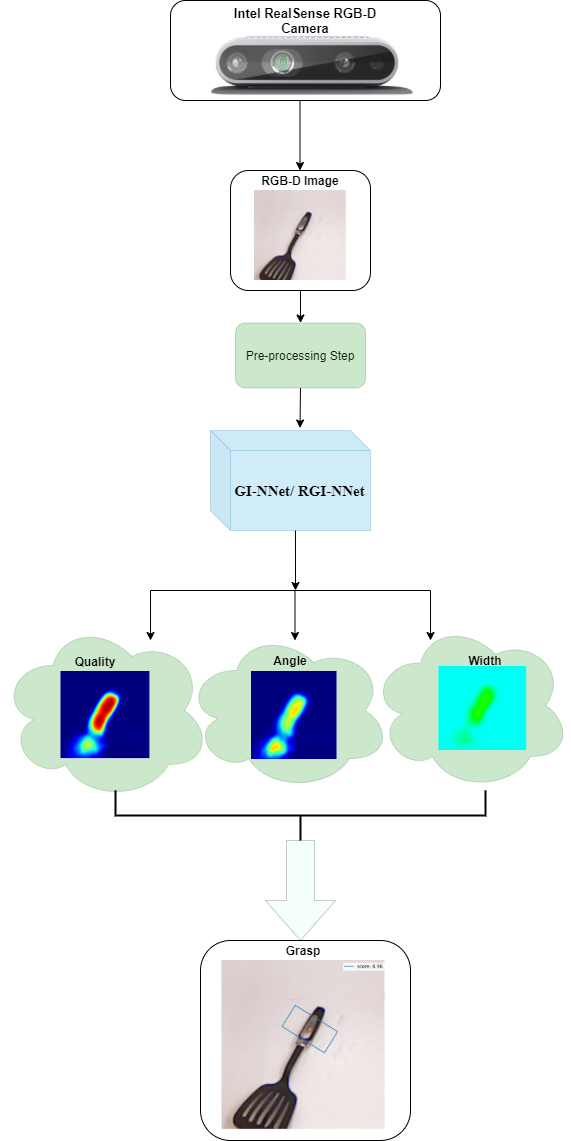}
\caption{Structural overview of our approach to predict an optimal grasp for an object.}
\label{fig:overview}
\end{figure}

In the present research we intend to address some of the problems associated with the intelligent grasping. More specifically, we have contributed in designing a lightweight and object independent model, GI-NNet which predicts grasps from the trained model and gives output as three sets of images. Our proposed model generates quality, angle and width images and grasps are inferred from these images at the  pixel level. This model is designed based on the concept of generative grasping  which is similarly proposed in Generative Grasping Convolutional Neural Network (GG-CNN) \cite{morrison2018closing} and Generative Residual Convolutional Neural Network (GR-ConvNet) \cite{kumra2019antipodal}. We have used Inception-Blocks which employ the similar idea as presented in Inception-V1 \cite{szegedy2015going} along with the ReLU activation function with learning rate of 0.001 in order to facilitate inference of information from a variety of kernel sizes as well as to keep the trainable parameters comparatively low (one third reduction) with significant improvement in accuracy (98.87\%) than the SOTA performing model \cite{kumra2019antipodal}.
But somehow the performance  of all the above mentioned models highly depends on the availability of labelled data. Owing to the scarcity of labelled data in the grasping domain, we subsequently propose to attach our model as decoder with a semi-supervised learning based architecture known as Vector Quantized Variational Auto Encoder (VQVAE), which we design to work efficiently when we train it with available labelled dataset as well as unlabelled data \cite{mridul}.  Our proposed GI-NNet integrated   VQVAE model, which we name as Representation based GI-NNet (RGI-NNet), has been trained with various splits of label data on CGD with as minimum as 10\% labelled dataset together with latent embedding generated from VQVAE up to 50\% labelled data with latent embedding obtained  from VQVAE. The performance level, in terms of grasp pose accuracy of RGI-NNet, varies between 92.13\% to 95.6\% which is far better than many other existing SOTA models trained with only labelled dataset.

Fig. \ref{fig:overview} illustrates the overview of our proposed model. We have trained our proposed models, GI-NNet and RGI-NNet over RGB-D and RGB images respectively to obtain corresponding grasps from generated Quality, Angle and Width images, which are then has been used to infer an optimal grasp at the pixel level. Following are the major contributions of the present research:
\begin{itemize}
    \item In this research, two novel grasp prediction models, GI-NNet and RGI-NNet,  have been designed to predict an optimal grasp at the pixel level. Subsequently, both the models have been trained and evaluated on CGD.
    \item Evaluation of GI-NNet on CGD provides promising increase in grasp accuracy of 98.87\% whereas GG-CNN \cite{morrison2018closing} and GR-ConvNet model \cite{kumra2019antipodal} reports SOTA success rate of 73.0\% and 97.7\% respectively.
    \item GI-NNet shows an improved success rate incorporating a lesser number of total trainable model parameters (592,300) as compared to GR-ConvNet \cite{kumra2019antipodal} (1,900,900). 
    \item The performance of proposed RGI-NNet architecture is analysed on CGD for split ratios of 0.1, 0.3, 0.5, 0.7, and 0.9 respectively. For minimal (10\%) labelled training data, it obtains an accuracy of 92.13\% which shows a significant performance on limited available grasping dataset.
    \item In the final output of the convolutional layers we have experimented with two transfer functions, Sigmoid and Tanh which provide  quality output and angle output (sin2$\Psi$ and cos2$\Psi$) respectively.
\end{itemize}

The rest of this paper is arranged in the following format: Section \ref{Relatedwork} discusses previous related research works with their limitations. In Section \ref{problemformulation} the problem formulation is elaborated. Section \ref{Preliminaries} depicts the grasp pose preliminaries with the concept of generative grasp approach, inception module, Variational Auto Encoder (VAE) and Vector Quantized Variational Auto Encoder (VQVAE) architectures, training dataset details and grasping metric formulation. Section \ref{methodology} describes detailed methodologies of our proposed model architecture along with the reasons for designing this model on rectifying the limitation of the previously proposed approaches. It also comprises the training method with incorporating loss and activation functions.
Section \ref{robotpose} illustrates the details about robotic grasp pose generation and execution.
Section \ref{result} present results and their comparative analyses. Conclusions and recommendations for the future research have been presented in the Section \ref{conclusion}.

\section{Related Work}
\label{Relatedwork}
In the field of robotics, intelligent robotic grasping has been one of the very compelling areas to the researchers. In spite of the fact that the issue appears to simply have the option to locate an appropriate grasp on any objects, the objective includes multifaceted components like different shapes and physical properties for the particular object.

Previously the robotic grasping research works are based on hand engineering \cite{maitin2010cloth}, \cite{kragic2003robust}, which is essentially time consuming but provided a headway in grasping utilizing multiple fingers \cite{kopicki2016one}, \cite{bohg2013data}. 
However, to acquire a steady grasp, the mechanics and contact kinematics of the end effector in contact with the object are examined from the study \cite{bicchi2000robotic}, \cite{shimoga1996robot}. 
Before advancing in machine learning, the grasping is performed by utilising the supervised learning and the models used to learn from synthetic data \cite{saxena2008robotic} with a restriction over the environments such as office, kitchen, and dishwasher. Overcoming these limitations require huge data and bigger models having much better learning/training  abilities which, these days, can be accomplished using deep learning architectures for detecting an skilled/optimal grasp. 

Fully Convolutional Grasp Quality Convolutional Neural Network (FC-GQ-CNN) predicts  an optimal grasp quality utilizing a data collection policy and synthetic training environment \cite{satish2019policy}. Though it generates stable grasps, recent research works are more focusing on RGB-D data to produce grasp poses. Various  research shows that deep neural network models are more useful for generating an efficient grasp pose for seen as well as unseen  objects \cite{schmidt2018grasping}, \cite{zeng2018robotic}. 

With the introduction of grasping rectangle concept in \cite{jiang2011efficient}, RGB-D images are utilized to detect a grasp pose with a two-step learning process which are as follows:
\begin{enumerate}
\item Primarily, the search space is narrowed down by sampling candidate grasp-rectangles.
\item Determining an optimal grasping-rectangle from the obtained candidate grasp-rectangles.
\end{enumerate}
A similar two-step method is used in \cite{lenz2015deep}, \cite{pinto2016supersizing} but the model performance decreases due to their large execution time. Whereas, AlexNet like architecture has also performed better on new objects by increasing the size of data \cite{pinto2016supersizing}. 
Due to large grasp inference time Morrison et al. introduces a generative approach, GG-CNN, that produces a grasp pose from depth images \cite{morrison2018closing}. Existing limitations of Computational complexity and discrete sampling are rectified through GG-CNN architecture. In a very recent research work, Kumra et al. \cite{kumra2019antipodal} has introduced a Generative Residual Convolutional Neural Network (GR-ConvNet), based on GG-CNN architecture, which predicts grasp more efficiently with a large number of trainable parameters. 
 However, the performances of all the models are highly dependent on the availability of labelled data which are not available in the sufficient quantity in the grasping domain and therefore, we subsequently propose to attach our model as decoder with a semi-supervised learning based architecture known as Vector Quantized Variational Auto Encoder (VQVAE), which we make to work efficiently when we train it with available labelled dataset as well as unlabelled data \cite{mridul}.  Our proposed GI-NNet integrated   VQVAE model, which we name as Representation based GI-NNet (RGI-NNet), has been trained with various splits of label data on CGD with as minimum as 10\% labelled dataset together with latent embedding generated from VQVAE up to 50\% labelled data with latent embedding obtained  from VQVAE. The performance level, in terms of grasp pose accuracy of RGI-NNet, varies between 92.13\% to 95.6\% which is far better than many other existing SOTA models trained with only labelled dataset.
The detailed architecture of our proposed models, GI-NNet and RGI-NNet, have been elaborated in the subsequent  sections.

\section{Problem Formulation}
\label{problemformulation}
Our research work is about to predict an optimal grasp pose with the proposed models from an image and execute it in a real time with a physical robot. 
In this work, robotic grasp pose is represented by (\ref{eqn:eq1}) 
where $G_R$ is the robot gripper grasp pose, $p_R=(x, y, z)$ is the center point position of the end-effector, $\Psi_R$ is gripper's rotation around $z$-axis, $w_R$ is the gripper's opening width and $q_R$ represents the effectiveness of the estimated grasp.
\begin{equation}
G_R = (p_R, \Psi_R, w_R, q_R)
\label{eqn:eq1}
\end{equation}

Let us consider an n-channel input image of size, $I = R ^{ n \times h \times w} $. On this image frame, the grasp can be defined by (\ref{eqn:eq2})
where, $s_I$ is the $(x,y)$ coordinate of the center point in pixels, $\Psi_I$ is the angle of rotation in camera frame, $w_I$ represents the width of grasp in pixel coordinates of image and $q_I$ is the effectiveness score of the grasp.
\begin{equation}
G_I = (s_I, \Psi_I, w_I, q_I)
\label{eqn:eq2}
\end{equation}

The effectiveness score, $q_I$ is the nature of the grasp at each point in the input image and is shown as a score esteem somewhere in the range of 0 and 1 where closeness to 1 denotes a more prominent possibility of a successful grasp irrespective of objects. 
$\Psi_I$ shows the antipodal estimation of angular rotation needed at each point to grasp the object of interest which has been addressed here in the range of $[$$-$$90\degree, 90\degree]$. 
$w_I$ represents the required width which has been kept in between $[0, w_{max}]$ pixels, where $w_{max}$ denotes the maximum width of the grasping rectangle.

For robotic grasp execution, predict grasp pose in image frame is used to infer the robotic grasp pose in robot's reference frame by applying a known set of transformations. The grasp pose relation between image frame and robot frame has been shown in (\ref{eqn:eq3}) where ${}^{C}T_{I}$ first transforms predicted grasp in an image frame to 3-D frame of a camera by means of its intrinsic parameters, then ${}^{R}T_{C}$ is used to transform camera frame coordinates to robot frame coordinates.
\begin{equation}
G_R={}^{R}T_{C}{}^{C}T_{I}(G_I)
\label{eqn:eq3}
\end{equation}
The set of grasps in image frame are then collectively represented by (\ref{eqn:eq4})
where $\Psi$, $W$ and $Q$ are each of $R^{h\times w} $ dimension and contain the values of angle, width and grasp quality respectively for each pixel in an image.
\begin{equation}
G=(\Psi, W, Q) \: \epsilon \: R^{3\times h\times w}
\label{eqn:eq4}
\end{equation}
To predict a grasp pose in an image frame we have designed models, GI-NNet and RGI-NNet, based on some preliminary ideas which have been discussed in the following section.

\section{Preliminaries}
\label{Preliminaries}
Primarily, in this work, a CNN and inception module based architecture, GI-NNet, has been proposed which deploys Inception Blocks to utilize the best kernel size for feature extraction \cite{Szegedy_2016_CVPR}. 
Further, we have investigated the Variational Auto Encoder to design the semi-supervised learning based model by integrating VQVAE and GI-NNet architectures to infer an optimal grasp rectangle with limited labelled data only.
Hence, it can be infer that our proposed models are also belong to the family of Generative grasp approaches. All the required concepts to design our models based on Generative grasp concept have been discussed in the following sections. 

\subsection{Generative grasp approach}
The approach of generative grasping is to generate a grasp pose for every pixel in the input image obtained from a camera. The input image thus obtained is passed through the generative convolutional neural network, which results in an output of four images. These four outputs are the grasp quality score, width of grasp, sine component of angle and the cosine component of the angle for every pixel of the input image. The two angle components are then processed to form a single angle output to infer the orientation of a grasp rectangle.
The advantage of this approach is to reduce the computational time compared to other SOTA approaches as discussed in \cite{morrison2018closing}. Our proposed approach for simultaneously generating grasp for an object with a lesser number of total trainable parameters.

\subsection{Inception Block}
\begin{figure}[t]
\includegraphics[scale=0.4]{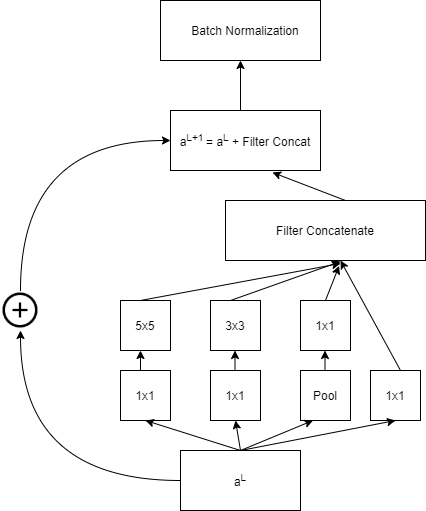}
\caption{Design of the Inception-Block architecture}
\label{fig:InceptionBlock}
\end{figure}

In this work, we have designed the inception block as a part of our proposed GI-NNet model in order to avoid any bias on the choice of kernel size. The detailed architecture of this inception module block is illustrated in Fig. \ref{fig:InceptionBlock}.
The structure of the inception block is inspired from \cite{szegedy2015going} and \cite{he2016deep}. We have used a minimal number of pools in order to minimize dependency on high value features. It has been observed that the inclusion of this block in the architecture produces significant results with comparatively lesser parameters.
The inception block after concatenating the output of different filters adds to the activation  of the previous layer to the concatenated output. This helps our architecture to incorporate properties of both inception and residual networks. The output of each inception block is then passed to a subsequent Batch Normalization layer to speed up the learning process.

\subsection{Variational Auto Encoder (VAE)}
For grasp detection, supervised learning approaches may be suitable in case of having a distribution of unlabelled training dataset with explicitly modelled latent representations \cite{mridul}. The probability density function for training samples over the latent variables (say, r) is defined as -
\begin{equation}
\mathbf{p}_{\theta}(\mathbf{x})=\int_{r} \mathbf{p}_{\theta}(\mathbf{r}) \mathbf{p}_{\theta}(\mathbf{x} \mid \mathbf{r}) \mathbf{d} \mathbf{r}
\label{eqn:pdf}
\end{equation}
After considering, a training sample is produced from the unexplored latent $r$ which is sampled over true priors of $p(r)$. Subsequently, the sample x is generated with the help of true priors and conditional Gaussian Distribution over the latent $r$. $p_{\theta}$ $(x|r)$ is designed to learn the optimal parameters which are obtained by maximizing the likelihood of the learning samples. Since the likelihood of samples is highly intractable, the posterior distribution of the model appears to be also intractable. To get rid of, true posterior approximation $q_{\phi}$ $(r|x)$ has been used which provides the lower bound of the training sample likelihood. Owing to the VAE architecture, the encoder is able to infers whereas the decoder help is generation. Both the encoder and decoder networks help in determining the mean and the diagonal covariance for the probabilistic density function.
The Evidence Lower Bound (ELBO) of the training samples likelihood can be determined as -
\begin{equation}
ELBO=\mathbf{E}_{\mathbf{r}}\left[\log \mathbf{p}_{\theta}(\mathbf{x} \mid \mathbf{r})\right]-\mathbf{D}_{\mathbf{KL}}\left(\mathbf{q}_{\phi}(\mathbf{r} \mid \mathbf{x}) \| \mathbf{p}_{\theta}(\mathbf{r})\right)
\label{eqn:ELBO}
\end{equation}

The encoder generated value enhances the likelihood of the learning samples subsequently the decoder produced value estimates the posterior which tends to the true prior. Hence, the VAE's encoder network empowers the inference of $q_{\phi}$ $(r|x)$ which can be further utilized during representation learning.

\subsection{Vector Quantized Variational Auto Encoder (VQVAE)}
In a research work, Vinyals et.al \cite{vinyals2015show} claims that the image can be modelled efficiently only utilising discrete symbols. It is being observed that when VAE architectures are designed with more efficient decoder which enables the ignorance in learning of latent vectors. This ignorance problem is termed as the posterior collapse. VAE architecture based model VQVAE solves this issue by keeping the discrete latents only. Along with this, latent embedding space is integrated to the base architecture of VAE. A latent embedding can be represented as - $e$ $\epsilon$ $ R^{N \times D}$. Here, $N$ denotes the number of embeddings comprise in latent embedding space and D denotes the embedding dimension. Encoder network generates $Z_e$ $(x)$ over the embedding vector space and following nearest neighbour lookup which outputs continuous vector that is being further quantized $Z_e$ $(x)$. The described process is termed as vector quantization. Later on, decoder performs reconstruction tasks with the quantized vectors.
The posterior distribution over the latent embedding is formulated as -
\begin{equation}
q(Z=k \mid x)=\left\{\begin{array}{ll}
1 & \text { for } \mathrm{k}=\operatorname{argmin}_{j}\left\|Z_{e}(x)-e_{j}\right\|_{2}, \\
0 & \text { otherwise }
\end{array}\right.
\label{eqn:post-dist}
\end{equation}

\subsection{Dataset}
For training purposes we have used the Cornell Grasping Dataset \cite{lenz2015deep} which consists of 885 RGB-D images of real objects comprising 5110 positive and 2909 negative grasps. The dataset represents antipodal grasps in  the form of rectangular bounding boxes with their coordinates being the pixels itself. We have augmented the data, since the used dataset is less in number, by means of random cropping, zooms and rotations resulting in nearly 51,000 grasps. We have experimented the model training procedure with augmentation and without augmentation. After examining the model performance in both the cases, it has been determined that training with augmentation helps in the learning process.

\subsection{Grasping Metric}
For the declaration of an optimal grasp we make use of the rectangle metric, proposed in \cite{jiang2011efficient}. This metric defines a generated grasp as a successful grasp if it satisfies the following two conditions:
\begin{enumerate}
    \item Intersection Over Union [IOU] score of the generated grasp rectangle and the ground truth grasp rectangle should be greater than 25\%.
    \item Offset in orientation of the grasp between generated and ground truth grasping rectangles should be less than 30$^{\circ}$.
\end{enumerate}

\begin{figure*}[t]
\includegraphics[width=\textwidth]{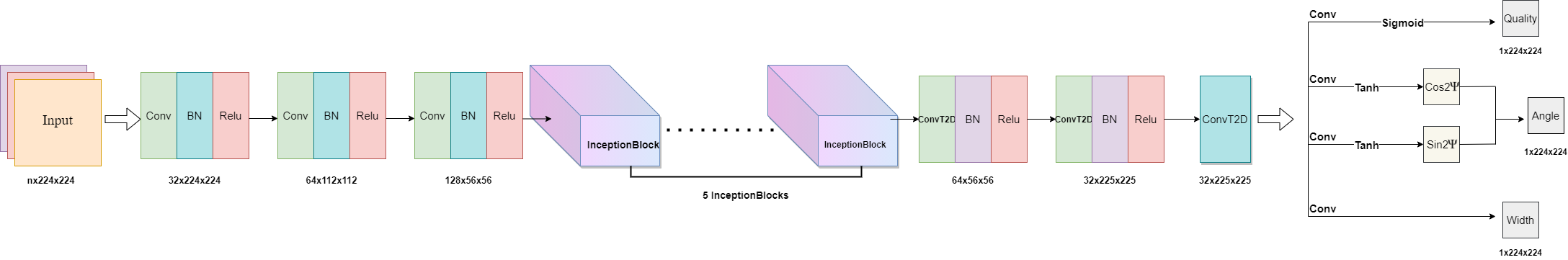}
\centering
\caption{Architecture of GI-NNet model}
\label{fig:Architecture}
\end{figure*}

\section{Methodology}
\label{methodology}
Initially, in the proposed approach to make an effective use of Inception Blocks to improve the efficiency of our inception block, we have tried incorporating ideas from Inception-V2 \cite{Szegedy_2016_CVPR}, but there is no significant improvement from it as those ideas are relevant to very deep networks.
Subsequently, we have experimented with dropout layers within different layers of the network to improve generalization capability of our proposed model, GI-NNet, and we found that incorporating three dropout layers, one before Inception Blocks, one within these blocks and one before feeding the network into the transposed convolutional layers, to get the best results.
Further, we have designed a cascaded model by integrating VQVAE and GI-NNet architectures to implement the semi-supervised learning approach.
Detailed architecture of our proposed models, GI-NNet and RGI-NNet with training details have been elaborated in following sections.  

\subsection{GI-NNet architecture} 
Our proposed GI-NNet requires RGB-D images as inputs and feeds them to a set of three 2-dimensional convolutional layers, then to five inception blocks followed by a set of transpose convolution layers and finally through a convolution operation, generating the desired output images, as illustrated in Fig. \ref{fig:Architecture}.

Output of the network consists of four images representing grasp quality, angle (sin2$\Psi$ and cos2$\Psi$) and grasp width, for each pixel of an image. These are generated after passing through transposed convolutional layers. We use such filter layers and padding to obtain the final output images of the same dimension as an input image. We make use of \emph{He} initialization (also known as Kaiming initialization) as proposed in \cite{he2015delving} which is suitable for usage when employing ReLU activation function, in-order to avoid exploding and vanishing gradient problems. This initialization sets the initialized weights to have a mean of 0 and standard deviation of $\sqrt{2/n}$, where n is the number of inputs to the node. 

\subsection{RGI-NNet architecture} 
\begin{figure*}[t]
\includegraphics[scale=0.5]{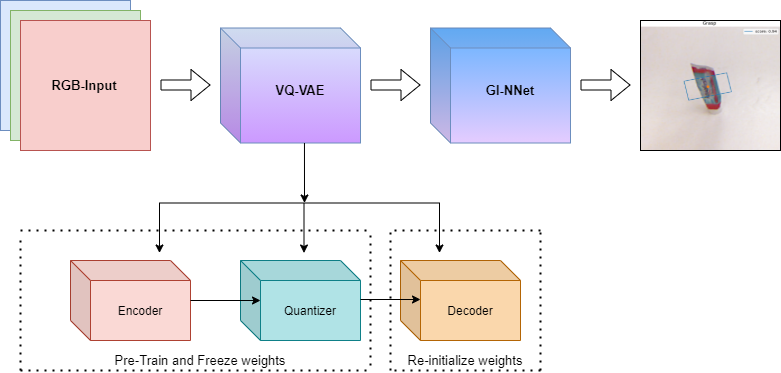}
\centering
\caption{Architecture of RGI-NNet model}
\label{fig:Cascaded_Architecture}
\end{figure*}

The proposed cascaded model takes RGB images as input and passes the input to the pre-trained VQVAE encoder which acts as a feature extractor and then a simple 2-D convolution operation is performed to match the dimensions of the latent space, the output is then passed to the Vector Quantization layers which translates the extracted features to discrete latent embeddings. The reinitialized  decoder then takes obtained latent embeddings and begins to reconstruct the images which are then fed to our proposed GI-NNet architecture. The re-initialization of the decoder helps the model to learn the reconstruction of the input image in a manner to get the best pixels for optimal grasp prediction. The detailed architecture of RGI-NNet is depicted in Fig. \ref{fig:Cascaded_Architecture}.

\subsection{GI-NNet training details} 
For training, 90\% data of CGD has been used and the remaining 10\% has been kept for evaluating our GI-NNet model performance. During training batch-size of 8 and the Adam optimizer have been used with a learning rate of 0.001. GI-NNet comprises a total 592,300 trainable parameters. It is observed that reduction in trainable parameters determines our model is inexpensive in terms of computation and faster in execution. Therefore, this lightweight feature suggests GI-NNet as a suitable model over closed-loop control applications in real-time robot grasp execution application.

\subsection{RGI-NNet training details}
The training approach employed here for the RGI-NNet model is to firstly train the VQVAE model to learn meaning-full latent embeddings for the inputted image. The entire CGD RGB images are fed in unlabelled form to train the VQVAE model. Once the VQVAE is trained we use its Encoder and Quantization layers to initialize our supervised learning based model. Subsequently, Decoder parameters are re-initialized for training the grasp prediction model. In this model, the grasp prediction model, GI-NNet, is trained with various labelled data splits such as 0.1, 0.3, 0.5, 0.7 and 0.9, here split represents fraction of labelled data used for training. The input is fed in batches of 8 to the respective layers of VQVAE and the Decoder of VQVAE then passes the output to GI-NNet which finally produces an optimal grasp. This models enables the grasp prediction with limited labelled training data only.
\subsection{Loss Function}
During experiments, Huber loss \cite{Huber2011} i.e the smooth L1 loss has been used which shows the best results, since it negotiates between absolute and square loss function. The loss is given by (\ref{eqn:loss}) and (\ref{eqn:zk}) where $\widehat{G}$ represents the predicted grasp and G represents the ground truth grasp.
\begin{equation} \label{eqn:loss}
L\left(G_{i}, \widehat{G}_{i}\right)=\frac{1}{n} \sum^{k} z_{k}
\end{equation}

\begin{equation} \label{eqn:zk}
z_{k}=\left\{\begin{array}{lc}
0.5\left(G_{i_{k}}-\widehat{G_{i_{k}}}\right)^{2}, & \text { if }\left|G_{i_{k}}-\widehat{G_{i_{k}}}\right|<1 \\
\left|G_{i_{k}}-\widehat{G_{i_{k}}}\right|-0.5 & \text { otherwise }
\end{array}\right.
\end{equation}

\subsection{Activation Function}
Initially, GI-Net is tested with Leaky ReLU, Parametric ReLU with gradient value of 0.1 and 0.001 respectively as the activation functions but it has been found ReLU activation with learning rate of 0.001 provides the most stable results.
So, with the convolutional layers we have used ReLU activation function, but in the final output layers Sigmoid and Tanh activation function have been used for quality image output and angle output image respectively. The grasp quality ground truth area is set to 1 and other pixels as 0. To improve these predictions we have used Sigmoid activation function. Similarly the sin and cos output images both are in range [-1, 1] so we have deployed Tanh activation function. For generating width output image, we have tried several activation functions but linear activation function turns out to be the most suitable one.

Obtained pose in image configuration space needs to convert into a robot configuration space for real time execution. So, it has been detailed in the following section.

\section{Robotic grasp pose generation and execution}
\label{robotpose}
The grasp generation performance of our proposed models, GI-NNet and RGI-NNet, are also being evaluated by executing on Anukul research robot in a real time experimental setup. To execute a grasp on physical robot, primarily generated grasp pose in an image space is mapped to the robot space. Then obtained grasp pose in robot space is used to get the robotic arm joint angles for the robotic grasp execution.

Initially, generated grasp rectangle in an image coordinate frame is obtained by using model GI-NNet or RGI-NNet. To execute it on a physical robot, the generated grasp pose is transformed from 2-D image frame to 3-D coordinates frame of camera and then followed by camera frame to robot frame transformation as shown in (\ref{eqn:eq3}). 
Primarily, external camera device has been calibrated and camera calibration matrix ($K$) has been achieved which has been shown in (\ref{eq:10}). Subsequently, obtained grasp pose in an image coordinates $(x, y)$ frame has been transformed to camera coordinates $(u, v, w)$ frame by using (\ref{eq:11}), (\ref{eq:12}) and (\ref{eq:13}) respectively.
\begin{equation}
K
=
\begin{bmatrix}
f_a & 0 & c_a \\
0 & f_b & c_b \\
0 & 0 & 1
\end{bmatrix}
\label{eq:10}
\end{equation}

\begin{equation}
    u = ((x - c_a) / (f_a)) \times depth[x][y] 
    \label{eq:11}
\end{equation}
\begin{equation}
    v = ((y - c_b) / (f_b)) \times depth[x][y]
    \label{eq:12}
\end{equation}
\begin{equation}
   w = depth[x][y] 
   \label{eq:13}
\end{equation}

The obtained grasp pose in camera coordinates is further mapped to robot configuration space by using ${}^{R}T_{C}$ which is obtained by a camera to robot mapping using \cite{hand-eye} and which finally calculates the joint angles for the generated robotic grasp pose ($G_{R}$) to execute the grasp.
All the details related to experimental setup and models' performance evaluation have been discussed in the subsequent sections.

\begin{figure}[t]
\includegraphics[angle=0, rotate=0,scale=0.045]{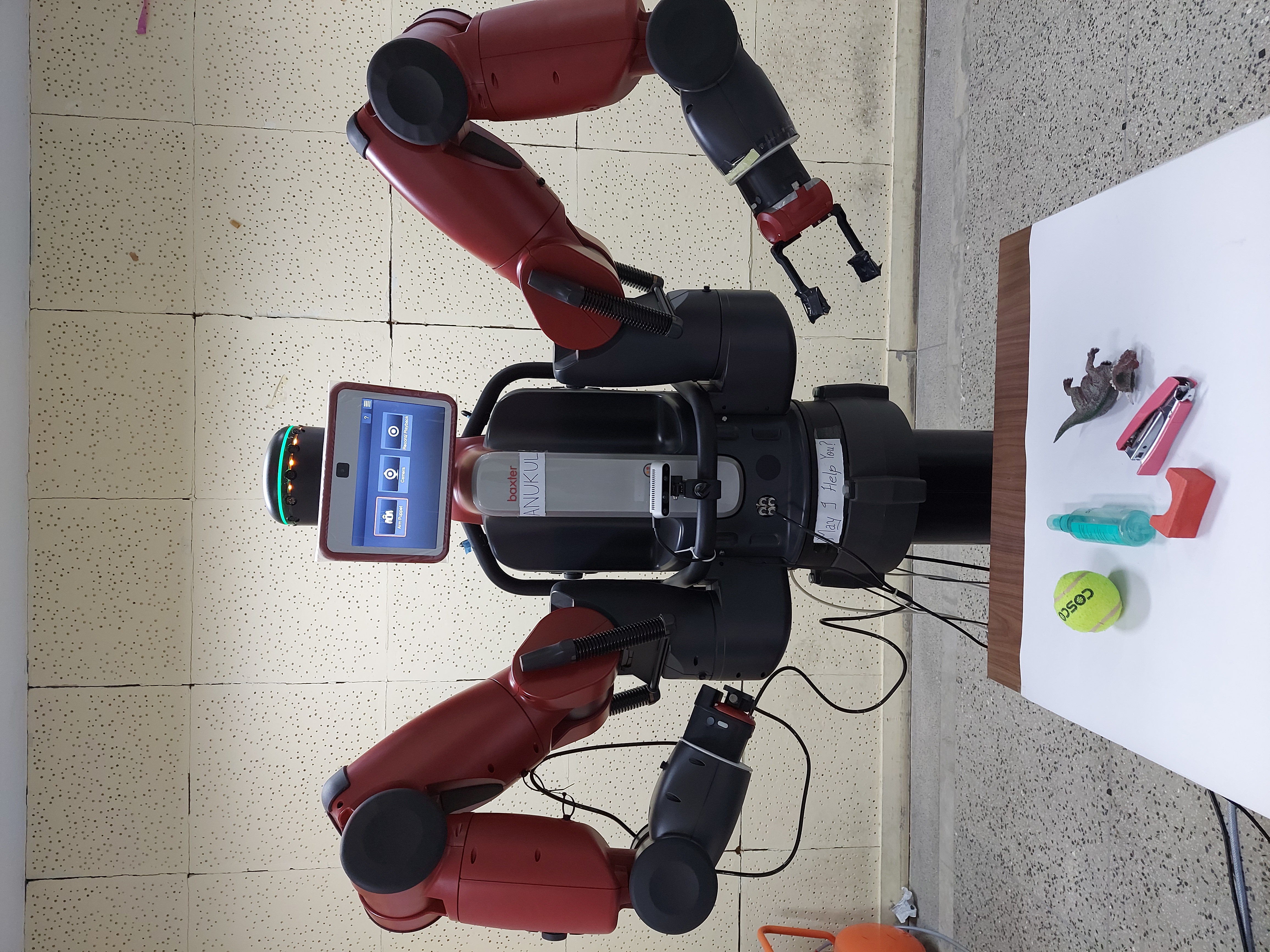}
\centering
\caption{Experimental setup}
\label{fig:setup}
\end{figure}

\begin{figure}[t]
\includegraphics[width=5cm,height=5 cm]{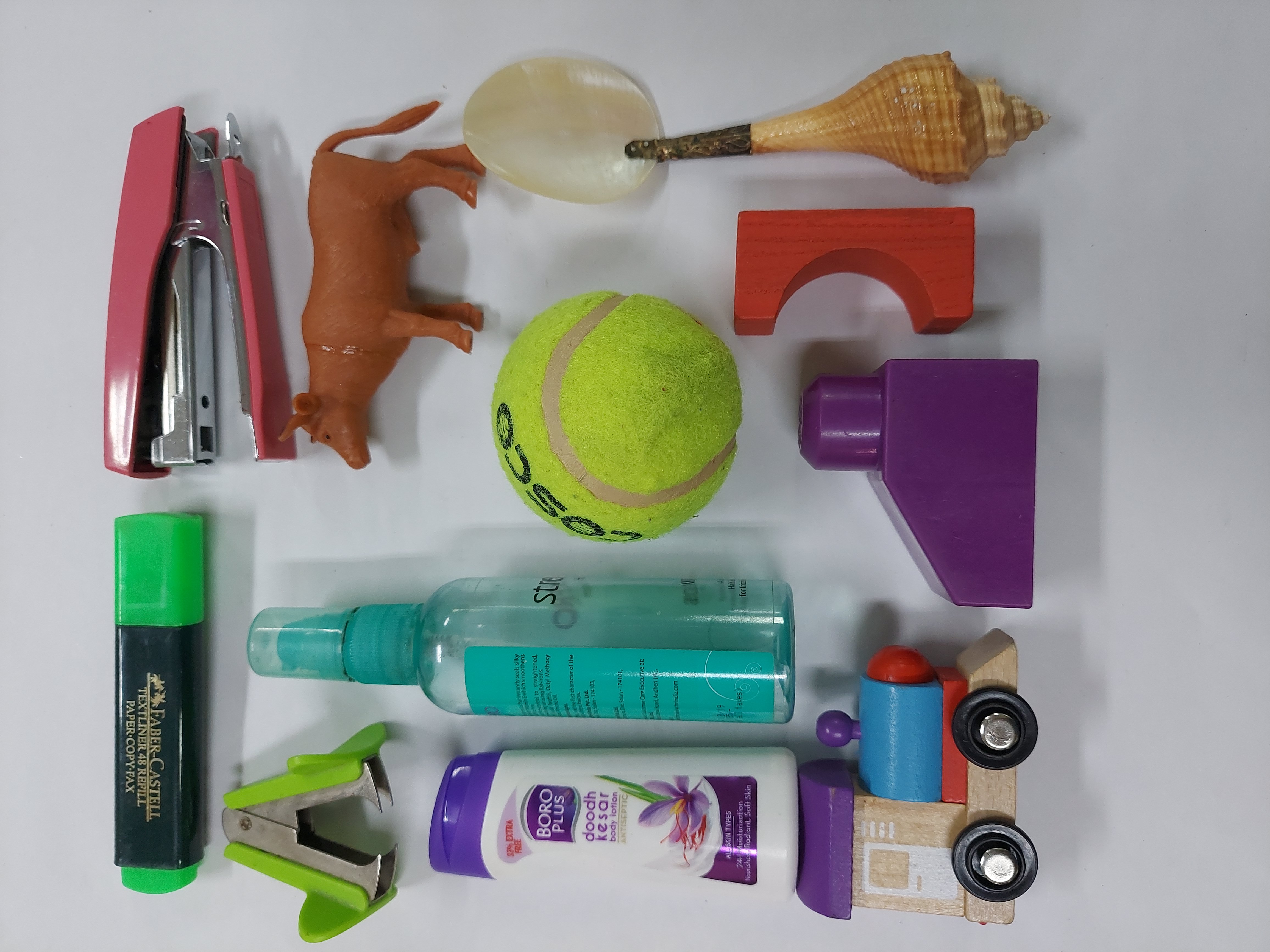}
\centering
\caption{Tested object's sample}
\label{fig:sampleobjects}
\end{figure}

\section{Experimental results and evaluation}
\label{result}
\subsection{Experimental Setup}
Anukul research robot hardware has been used for verification of the grasp pose prediction with robotic grasp execution. The setup for the experiments in a real-time environment has been shown in Fig. \ref{fig:setup}. Anukul has three internal cameras, two on both the arm's EE and one on top of the head with maximum resolution of $1280\times800$ pixels incorporating 30 Frame Per Second (FPS) and focal length of 1.2 mm. However for our research for grasp manipulation,  object depth is required and hence we had to  mount  an additional camera on the robot torso, with much higher resolutions (\cite{D435}) with appropriate calibrations so as to get maximum work space visibility . We have used Baxter Software Development Kit (SDK) which has been developed in Robot Operating System (ROS) environment for computation of inverse kinematics, trajectory planning, with appropriate speed control using Python language.
Though the robot model has both electrical and vacuum grippers, for the experiments the left arm with an electrical gripper has been used,although without loss of generality the other arm and the other grippers could also be used. For real time grasping candidates, Regular/Irregular shaped medium-sized objects have been used. Some sample objects have been presented in Fig. \ref{fig:sampleobjects}.
\begin{figure}
\centering
\includegraphics[scale=0.35]{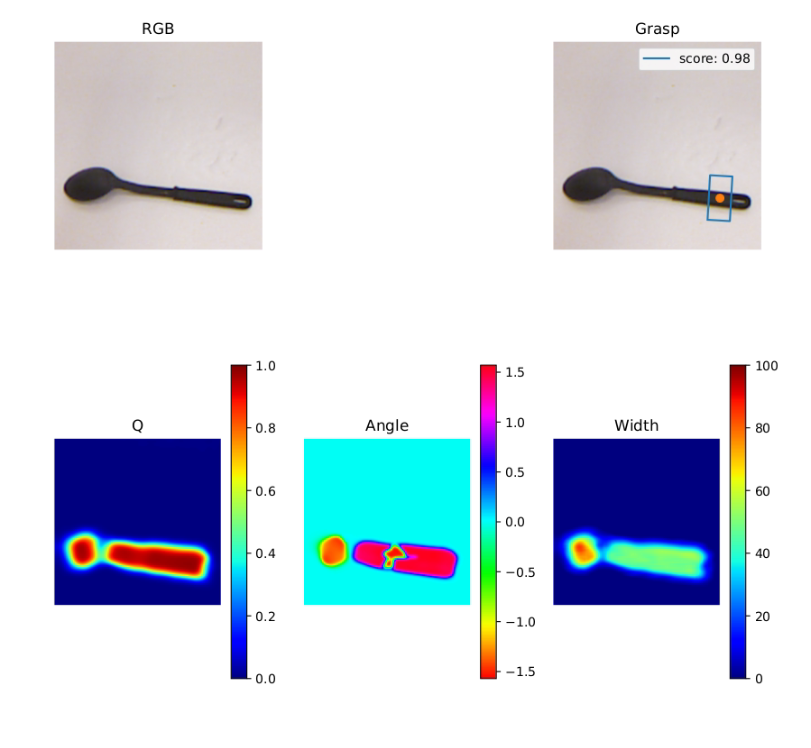}
\caption{Pose prediction with GI-NNet model}
\label{fig:example}
\end{figure}

\begin{figure*}[t]
\includegraphics[width=.8\textwidth]{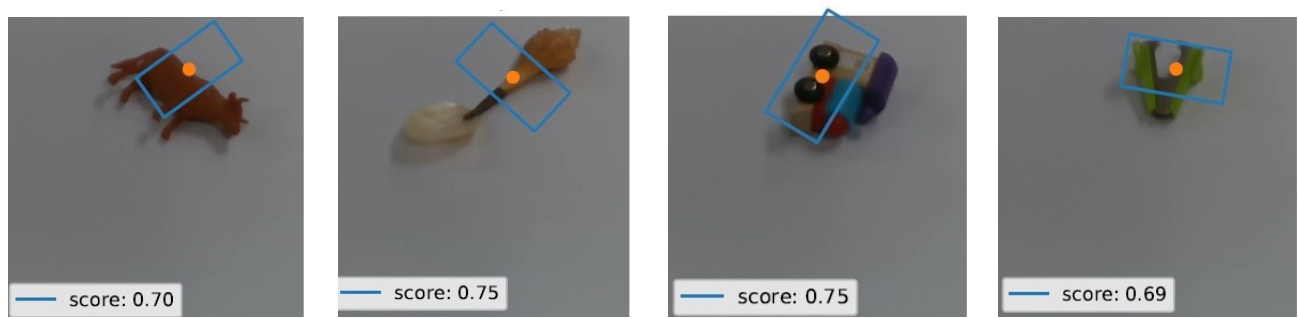}
\centering
\caption{Performance of GI-NNet on unseen object images}
\label{fig:unseen obj}
\end{figure*}

\subsection{GI-NNet model evaluation}
To validate the proposed model we have tested and compared it with the SOTA approaches along with the  performance analyses on CGD. 
The  proposed GI-NNet model is found to be robust, applicable for real time environment, and it generalizes well  towards various geometrical shapes and sizes  of the objects and thus making it suitable for the closed loop control of grasp execution. 
To test the generalization capability of the proposed model, GI-NNet is evaluated on the CGD which predicts optimal grasp pose on object images with the  SOTA success rate of 98.87\%. The output of GI-NNet is shown for an object image in Fig. \ref{fig:example}. Here, the image frame grasp $G_I$ is generated through the optimal grasp quality $q_I$ image, the grasp angle for the grasping $\Psi_I$ image and the grasp width $w_I$ image. The bounding rectangle on the object represents the predicted grasp pose.
Apart from testing on CGD images, we evaluated the proposed model on some unseen object images which are further illustrated in Fig. \ref{fig:unseen obj}.
Moreover, it attains such performance incorporating total trainable model parameters of 592,300 which is much lesser in comparison to the other related models which have been discussed in upcoming subsections.

\subsection{RGI-NNet model evaluation}
RGI-NNet is trained on CGD with split percentage of 10, 30, 50, 70 and 90, and the performance of each trained model has been presented in the Table \ref{tab:Table-RGI-NNet}. 
\begin{table}[!ht]
\centering
 \begin{tabular}{| c | c | c |} 
 \hline
 \textbf{Split Ratio} & \textbf{Mahajan et.al.\cite{mridul}} & \textbf{Ours (RGI-NNet)}  \\ 
 \hline
 0.1 & 85.3933 &92.1348  \\\hline 
 0.3 & 87.6404 &95.5056  \\\hline
 0.5 & 89.8876 &95.5056  \\\hline
 0.7 & 89.8876 &96.6292  \\\hline
 0.9 & 89.8876 &95.5056  \\\hline 
 \end{tabular}
 \caption{Performance comparison with different split ratio}
 \label{tab:Table-RGI-NNet}
\end{table}
which clearly shows that the model provides stable grasping performance though trained on very limited amount of labelled data. Moreover, it attains a significant accuracy of 92.1348\% with 10\% of labelled data only. Our proposed model, RGI-NNet, is able to achieve such significant improvements in results due to the more powerful decoder training with our proposed, GI-NNet, model.  Apart from this, we have also verified its performance for generating an optimal grasp on seen as well as unseen objects.
\subsection{Comparative studies with the existing SOTA approaches}
The cross-validation approach as mentioned in various research works \cite{lenz2015deep}, \cite{redmon2015real}, \cite{kumra2019antipodal} and \cite{asif2018ensemblenet}, we also have tried to follow the similar approach of image-based data splits. Our model has achieved an accuracy of 98.87\% on image-based split as depicted in the Table \ref{tab:Table-1}, which compares the success rate of our model  with that of the  SOTA models. 
GI-NNet outperforms other enlisted approaches as mentioned in the Table \ref{tab:Table-1}.  Our inception module based network predicts optimal grasps for  various kinds of geometrically distinct objects from the validation set.

\begin{table}[!ht]
\centering
 \begin{tabular}{| c | c |} 
 \hline
 \textbf{Model} & \textbf{Accuracy (\%)}  \\ 
 \hline
 Fast Search \cite{jiang2011efficient} & 60.5  \\\hline 
 GG-CNN \cite{morrison2018closing} & 73.0  \\\hline
 SAE, struct. reg. \cite{lenz2015deep} & 73.9  \\\hline
 Two-stage closed-loop \cite{wang2016robot} & 85.3  \\\hline
 AlexNet, MultiGrasp \cite{redmon2015real} & 88.0  \\\hline 
 STEM-CaRFs \cite{asif2018ensemblenet} & 88.2  \\\hline 
 GRPN \cite{karaoguz2019object} & 88.7  \\\hline 
 ResNet-50x2 \cite{kumra2017robotic} & 89.2  \\\hline 
 GraspNet \cite{asif2018graspnet} & 90.2 \\\hline 
 ZF-net \cite{guo2017hybrid} & 93.2  \\\hline 
 FCGN, ResNet-101 \cite{zhou2018fully} & 97.7  \\\hline 
 GR-ConvNet \cite{kumra2019antipodal} & 97.7  \\\hline 
 \textbf{GI-NNet}  & \textbf{98.87}  \\\hline 
 \end{tabular}
 \caption{Performance comparison}
 \label{tab:Table-1}
\end{table}

Utilizing the data augmentation approach during the training of the network on the CGD the success rate in generating grasp poses of the GI-NNet model is improved. Moreover, such an improved performance has been observed with much reduced trainable parameters as mentioned in the Table \ref{tab:Table-2}.
\begin{table}[!ht]
\centering
 \begin{tabular}{| c | c |} 
 \hline
 \textbf{Model} & \textbf{Parameters}  \\ 
 \hline
 GR-ConvNet  &  1,900,900  \\\hline 
 GI-NNet  & 592,300 \\\hline
 \end{tabular}
 \caption{Comparison of model parameters}
 \label{tab:Table-2}
\end{table}

\begin{figure*}[hbt!]
\includegraphics[width=0.7\textwidth]{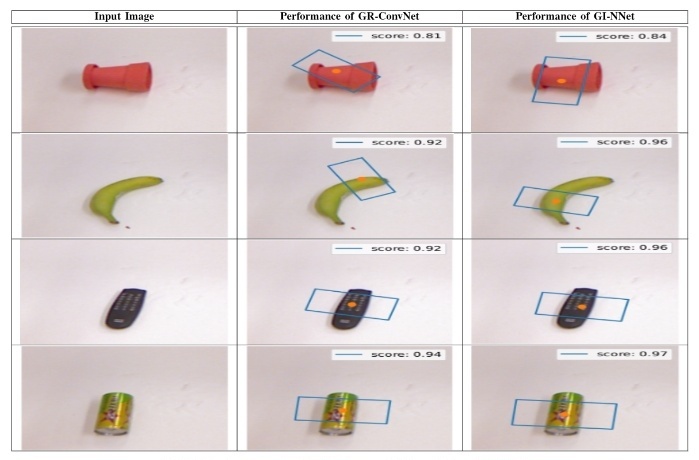}
\centering
\caption{Performance comparison between GR-ConvNet and GI-NNet}
\label{fig:CompAnalysis}
\end{figure*}

 Such drastic reduction in trainable parameters makes the GI-NNet far more efficient, fast and computationally inexpensive. Thus this model can be deployed in closed loop control on a robot grasp execution in real time applications. We also make a comparative analysis on the performance of GR-ConvNet and GI-NNet models on predicting an optimal grasp for object images from CGD as shown in the Fig. \ref{fig:CompAnalysis}. 
Subsequently, the performance of RGI-NNet has been compared with Mahajan et.al.\cite{mridul} and it is being observed that the performance is highly improved with limited labelled data. 
 From the above mentioned studies, we can claim that GI-NNet is  comparatively a much better model to generate optimal grasp pose for the intelligent robot grasping applications when we have sufficient labelled data for training, whereas RGI-NNet has the ability to predict optimal grasp pose when we have limited labelled data for training, making it an attractive model for intelligent robot grasping.
 

\section{Conclusions and recommendations for future research}
\label{conclusion}
In the present investigation, we have presented two new grasping models, GI-NNet and RGI-NNet, where GI-NNet is capable of generating optimal grasp rectangle in a superior way at the pixel level for the object in a scene with sufficient labelled data and RGI-NNet model is able to predict an optimal grasp with limited labelled data. Both the models are fast enough to deploy in a closed loop robot grasp execution control owing to much better performance for the grasp inference. During experiments, we have achieved an improvement in predicting grasp poses with more efficiency for seen as well as unseen objects. 
With rigorous experimentation we have confirmed that when there is sufficient labelled data for training then the GI-NNet model may perform better  but when there is limited labelled data for training then the performance of our proposed RGI-NNet model is much better, perhaps due to the inherent capability of VQVAE of generating data suitable for training the network parameters .
Currently, models are trained and tested only on CGD images but the same models can be trained on other available labelled dataset also to check for the performance in future.
We can also use our proposed models to bring agility by creating sufficient labelled data for novel objects’ category which could make the training more efficient. 
In the present investigation we have used only table top grasping for experimental verification of grasp efficiency which may be generalised with  more realistic grasping by analysing 6-D pose estimation.

\bibliographystyle{ieeetr}
\bibliography{bibliography.bib}

\end{document}